# Addressing the Memory Bottleneck in AI Model Training

David Ojika[*], Bhavesh Patel[^], G. Anthony Reina[†], Trent Boyer[†], Chad Martin[†], Prashant Shah[†]
[*]University of Florida
[^]Dell EMC
[†]Intel

## ABSTRACT

Using medical imaging as case-study, we demonstrate how Intel-optimized TensorFlow on an x86-based server equipped with 2nd Generation Intel Xeon Scalable Processors with large system memory allows for the training of memory-intensive AI/deep-learning models in a scale-up server configuration. We believe our work represents the first training of a deep neural network having large memory footprint (~ 1 TB) on a single-node server. We recommend this configuration to scientists and researchers who wish to develop large, state-of-the-art AI models but are currently limited by memory. A poster of this work presented at Workshop on MLOps Systems has been made available at https://bit.ly/2W6EwGf.

## 1  Introduction

Medical image analytics, such as semantic segmentation, are particularly challenging because the segmentation model is trained to automatically classify individual voxels from large volumetric images [1]. The 3D (and sometimes 4D) nature of this data type demands increased memory capacity and processing power when training the model. Consequently, researchers resort to tricks, such as downsizing and tiling images, to cope with available system memory or adopting shallower neural network topologies to address the high processing requirement. Ultimately, most researchers choose a model based on the memory limitations of the hardware rather than based on the best possible model design.

A high-memory CPU-based server solution (e.g., 2nd Generation Intel Xeon Scalable Processors), present an attractive architecture for addressing the compute and memory requirement of 3D semantic segmentation algorithms, such as 3D U-Net model. With more than 1 TB of system memory available, the 2nd Generation Intel Xeon Scalable Processor allows researchers to develop large deep learning (DL) models that can be several orders of magnitude larger than those available on DL accelerators.

## 2  Multi-Modal Brain Tumor Analysis

Multimodal brain tumor analysis is an important diagnosis process in medical imaging. A brain tumor occurs when abnormal cells form within the brain. Gliomas are the most frequent primary brain tumors in adults, presumably originating from glial cells and infiltrating the surrounding tissues [2]. Current imaging techniques used in clinical studies are limited to basic assessments, indicating for example, the presence of gliomas, or limited to non-wholistic coverage of the scan as a result of the reliance on rudimentary measurement techniques [3]. By replacing current assessments with highly accurate and reproducible measurements, AI and DL techniques can automatically analyze brain tumor scans, providing an enormous potential for improved diagnosis, treatment planning and patient follow-ups.

A typical MRI scans of the brain may contain 4D volumes with multimodal, multisite MRI data (FLAIR, T1w, T1gd, T2w). With appropriate training data sets, an AI-based brain tumor analysis solution should perform segmentation on the images, annotating regions of interest as necrotic/active tumor, oedema or benign.

### 2.1  Computing Challenges

While the high *processing* requirement of medical data analysis may be addressed with hardware accelerators, such as GPUs, addressing the *memory* requirement is not straightforward. As an example, a GPU accelerator has between 8 GB to 32 GB of memory. Although convolutional neural networks may only have several million trainable parameters, the actual memory footprint of these models is not due to solely those parameters. Instead, most of the memory footprint of these models comes from the activation (feature) maps in the model. These activation maps—essentially copies of the original images—are a function of the size of the input to the network. Therefore, models that use large batch, high resolution, high dimensional image inputs often require more memory than the accelerator card can accommodate. As a simple example, a ResNet-50 topology that can train successfully on a 224x224x3 RGB input image may report an out of memory (OOM) error when training on 4096x2160x3 input images common to 4k video streams. To compensate for the memory constraints of accelerator cards, researchers use "tricks" such as image/batch resize, tiling/patching, model compression, model parallelism.

Although these tricks have been used to produce clinically-relevant models, we believe that researchers would not choose to use them if it were not for the memory limitations in hardware. In other words, these tricks were not created to obtain better models—they are instead necessary workarounds for hardware limitations.

### 2.2  3D U-Net Model

Convolutional neural networks (CNNs) such as U-Net have been widely successfully in 2D segmentation in computer vision problems [6]. However, most medical data used in clinical practice consists of 3D volumes. Since only 2D slices can be displayed on a computer screen, annotating these large volumes with segmentation labels in a slice-by-slice manner is cumbersome and inefficient. 3D U-Net [7], based on U-Net architecture, performs volumetric segmentation by taking 3D volumes as input and processing them with corresponding 3D operations: 3D convolutions, 3D max-pooling, 3D up-sampling, etc. The resulting output is a trained model that reasonably generalizes well since the image slices contain mostly repetitive structures with corresponding variation. In general, the 3D U-Net model is both computation- and memory-intensive.

## 2.3 Experimental Data

The medical decathlon dataset [4] is a 3D semantic segmentation challenge with a broad range of medical imaging tasks including tumor and cancer diagnoses for various parts of the human body, including the liver, brain, lung, colon, and prostate. The images were generated either through a CT or an MRI scan at various universities and research centers from across the globe. Given this variety of data, the images present the opportunity for data scientists and machine learning practitioners to optimize AI algorithms for generalizability in medical imaging tasks with a primary focus on semantic segmentation. Thus, the most commonly used metric in segmentation tasks, Dice Similarity Coefficient (DSC) [5], along with Normalized Surface Distance (NSD) (distance between reconstructed surfaces) are used to assess different aspects of the performance of each task and region of interest. In this paper, we focus on the DSC (or simply, "dice coefficient") of the Brain Tumor task from the BraTS dataset, which contains 750 4D MRI volumes: 484 for training and 266 for testing.

## 3. Results

A single-node server with large memory has the potential to reduce organizations' total cost of ownership (TCO), while addressing the memory bottleneck involved with training large models with complex datasets. Using a 4-socket 2nd Generation Intel Xeon Scalable Processor system on a Dell EMC PowerEdge server equipped with 1.5 TB of system memory we trained the 3D U-Net model with the BraTS dataset (using only the "FLAIR" channel) without the need for scaling down the data nor tiling images to fit in memory. We used Intel-optimized TensorFlow - available as an Anaconda library [9] - and Conda as the Python virtual execution environment. The Intel-optimized TensorFlow distribution incorporates Deep Neural Network Library (DNNL) [10] (formerly MKL-DNN), allowing us to leverage the processors' underlying hardware features, including high CPU core count (80 cores), AVX-512 for floating-point operations, and integrated memory controllers supporting 1TB-per-socket system memory, to speed up the training process.

Using this system configuration, we achieved, within 25 training iterations (epochs), close to state-of-the-art performance: 0.997 accuracy, 0.125 loss and 0.82 dice coefficient. We also profiled the memory footprint of the training task, comparing the results (Fig 1) with our theoretical calculations and found our estimations to be accurate for our chosen hyperparameters (batch, feature-map, and image sizes). Meanwhile, the training speed (TS) for a single step (involving forward pass and backward pass of a single 3D scan) per training epoch was 30 seconds per image, a 3.4x speedup (Fig 2) compared to stock TensorFlow (without DNNL) at the same training batch size of 16.

Fig 3 depicts the prediction performance of the trained model. As observed, the segmentation mask from the model predictions closely match the ground truth mask. Using the TS and epoch count as a reference, machine learning practitioners can "plug in" their specific training data and hyperparameters to estimate both the required system memory and task completion time when training their own deep learning models on Intel architecture.

## 4. Conclusions

In this paper, we presented the multimodal brain tumor analysis for medical diagnosis, highlighted the computing challenges, and presented the 3D U-Net model for the task of volumetric image segmentation. With a memory-rich sever having 1.5 TB system memory, we trained the 3D U-Net model using the BraTS dataset (a medical segmentation benchmark) and achieved close to state-of-the-art accuracy of 0.997 and dice coefficient of 0.83. To the best of our knowledge, the results presented in this paper represent the first milestone in training a deep neural network having large memory footprint (close to 1 TB) on a single-node server without hardware accelerators like GPUs. Further, by enabling Deep Neural Network Library (DNNL) optimizations, we achieved a speedup of 3.4x per training step compared to stock TensorFlow. Future work will involve replicating the single-node, memory-rich experimental setup described in this paper into a multi-node CPU cluster, where we can expect to see greatly enhanced training performance of the 3D U-Net model and potentially that of other complex 3D models and datasets.

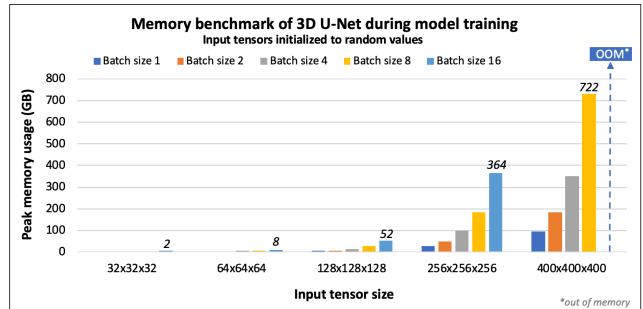

Fig 1. Figure 3. Benchmarking the memory usage of 3D U-Net model-training over various input tensors

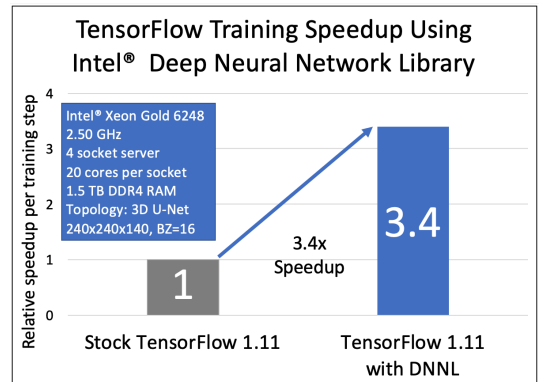

Fig 2. TensorFlow with Deep Neural Network Library (DNNL) enabled achieves increased performance versus stock TensorFlow (without DNNL).

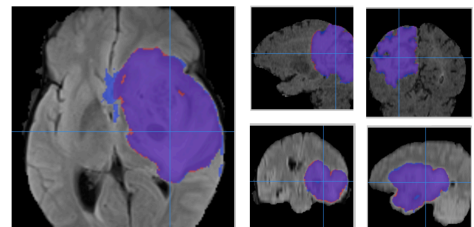

Fig 3. Prediction performance of the trained model, showing a slice of the brain from different views. The red overlay is the prediction from the model and the blue overlay is the ground truth mask. Any purple voxels are true positives.

# REFERENCES

[1] Swarnendu Ghosh, Nibaran Das, Ishita Das, and Ujjwal Maulik. 2019. Understanding Deep Learning Techniques for Image Segmentation. *ACM Comput. Surv.* 52, 4, Article 73 (August 2019), 35 pages.

[2] Holland, E n.d., 'Progenitor cells and glioma formation', Current Opinion in Neurology, vol. 14, no. 6, pp. 683–688.

[3] B. H. Menze *et al*., "The Multimodal Brain Tumor Image Segmentation Benchmark (BRATS)," in *IEEE Transactions on Medical Imaging*, vol. 34, no. 10, pp. 1993-2024, Oct. 2015.

[4] BRATS dataset https://www.med.upenn.edu/sbia/brats2018.html

[5] Maier-Hein, Lena, Eisenmann, Matthias, Reinke, Annika, Onogur, Sinan, Stankovic, Marko, Scholz, Patrick, & Full, Peter M. (2018). Is the winner really the best? A critical analysis of common research practice in biomedical image analysis competitions (Version 1.0.0) Zenodo.

[6] Olaf Ronneberger, Philipp Fischer & Thomas Brox. U-Net: Convolutional Networks for Biomedical Image Segmentation. Medical Image Computing and Computer-Assisted Intervention (MICCAI), Springer, LNCS, Vol.9351, 234--241, 2015

[7] Ciçek Ö., Abdulkadir A., Lienkamp S.S., Brox T., Ronneberger O. (2016) 3D U-Net: Learning Dense Volumetric Segmentation from Sparse Annotation. In: Ourselin S., Joskowicz L., Sabuncu M., Unal G., Wells W. (eds) Medical Image Computing and Computer-Assisted Intervention – MICCAI 2016. MICCAI 2016. Lecture Notes in Computer Science, vol 9901. Springer, Cham

[8] H. R. Roth, C. Shen, H. Oda, M. Oda, Y. Hayashi, K. Misawa, and K. Mori, "Deep learning and its application to medical image segmentation," Medical Imaging Technology, vol. 36, no. 2, pp. 63–71, 2018.

[9] Intel Optimization for TensorFlow. https://software.intel.com/en-us/articles/intel-optimization-for-tensorflow-installation-guide

[10] Deep Neural Network Library. https://intel.github.io/mkl-dnn

[11] Bakas, S., Akbari, H., Sotiras, A., Bilello, M., Rozycki, M., Kirby, J.S., Freymann, J.B., Farahani, K., & Davatzikos, C. (2017). Advancing The Cancer Genome Atlas glioma MRI collections with expert segmentation labels and radiomic features. *Scientific data*.

[12] Spyridon Bakas, Hamed Akbari, Aristeidis Sotiras, Michel Bilello, Martin Rozycki, Justin Kirby, John Freymann, Keyvan Farahani, and Christos Davatzikos. (2017) Segmentation Labels and Radiomic Features for the Pre-operative Scans of the TCGA-GBM collection. The Cancer Imaging Archive.

[13] Using Intel® Xeon® for Multi-node Scaling of TensorFlow with Horovod. https://www.intel.ai/using-intel-xeon-for-multi-node-scaling-of-tensorflow-with-horovod/#gs.mqqqpc